\documentclass[letterpaper, 10pt, conference]{ieeeconf}      

\IEEEoverridecommandlockouts 
\usepackage{cite}
\usepackage{amsmath,amssymb,amsfonts}
\usepackage{graphicx}
\usepackage{textcomp}
\usepackage{amsmath}

\DeclareMathOperator*{\argmin}{arg\,min}
\usepackage[table,xcdraw]{xcolor}
\usepackage{authblk}
\let\labelindent\relax
\usepackage{enumitem}
\usepackage{hyperref}
\usepackage{breqn}
\usepackage{algpseudocode}
\usepackage[linesnumbered,ruled,lined]{algorithm2e}
\usepackage{subfigure}

\usepackage{array}
\usepackage{makecell}

\allowdisplaybreaks

\title{\LARGE \bf
Bi-Level Optimization Augmented with Conditional Variational Autoencoder for Autonomous Driving in Dense Traffic
}
\author{Arun Kumar Singh$^{1}$, Jatan Shrestha$^{1}$, Nicola Albarella$^{2}$%
\thanks{$^{1}$University of Tartu, Estonia \quad $^{2}$University of Naples Federico II, Italy}%
\thanks{All authors contributed equally; ordering is purely alphabetical. Code is available on the project website: \url{https://sites.google.com/view/mpc-bi-level}}%
}

\begin{document}
\maketitle

\begin{abstract}
Autonomous driving has a natural bi-level structure. The goal of the upper behavioural layer is to provide appropriate lane change, speeding up, and braking decisions to optimize a given driving task. However, this layer can only indirectly influence the driving efficiency through the lower-level trajectory planner, which takes in the behavioural inputs to produce motion commands. Existing sampling-based approaches do not fully exploit the strong coupling between the behavioural and planning layer. On the other hand, end-to-end Reinforcement Learning (RL) can learn a behavioural layer while incorporating feedback from the lower-level planner. However, purely data-driven approaches often fail in safety metrics in unseen environments. This paper presents a novel alternative; a parameterized bi-level optimization that jointly computes the optimal behavioural decisions and the resulting downstream trajectory. Our approach runs in real-time using a custom GPU-accelerated batch optimizer, and a Conditional Variational Autoencoder learnt warm-start strategy. Extensive simulations show that our approach outperforms state-of-the-art model predictive control and  RL approaches in terms of collision rate while being competitive in driving efficiency.


\end{abstract}


\section{Introduction}
Motion planning for autonomous driving can be divided into two hierarchical components. At the top level, the behavioural layer computes decisions such as lane change, speeding up and braking based on the traffic scenario and the driving task. The behavioural inputs can be conveniently parameterized as set-points for longitudinal velocity, lateral offsets from the centre line, goal positions, etc. The advantage of such representation is that it naturally integrates with the lower-level optimal trajectory planner \cite{hoel_rl_behavior}, \cite{rl_behavior_connected}, \cite{huegle2019dynamic}, \cite{li2021safe} \cite{wei2014behavioral}, \cite{lim2019hybrid}. The behavioural layer can be critical for driving in dense traffic as it can guide the lower-level planner into favourable state-space regions, much in the same way a collision-free global plan can make the task of the local planner easier.

\noindent \subsubsection*{Existing Gaps} The aim of the behavioural layer is to come up with correct lane change, acceleration and braking decisions to optimize a given driving task. However, it can only indirectly affect the driving efficiency through the lower-level trajectory planner. Clearly, there is strong inter-dependency between the two layers. However, existing trajectory sampling approaches \cite{li2021safe} \cite{wei2014behavioral} do not fully exploit this inter-dependency (see Fig. \ref{pipeline_fig}(a) and Section \ref{connections}). For example, these works have no mechanism for modifying the behavioural inputs based on how the associated lower-level trajectory performs on the given driving task. Reinforcement learning (RL) techniques address this drawback by learning the behavioural layer. The rewards from the environment act as a feedback to modify the behavioural inputs while taking into account the effect of the lower-level planner \cite{hoel_rl_behavior}, \cite{rl_behavior_connected}, \cite{huegle2019dynamic}, \cite{li2021safe}. Though effective, especially in sparse traffic scenarios, the purely data-driven approaches typically struggle with safety metrics in unseen environments.


In this paper, we present a novel approach that estimates the direction in which the behavioural inputs need to be perturbed in order to improve the optimality of the lower-level trajectory with respect to the driving task (Fig.\ref{pipeline_fig} (a)). Our core contributions are:



\begin{figure}[!t]
    \centering
    \includegraphics[scale = 0.34]{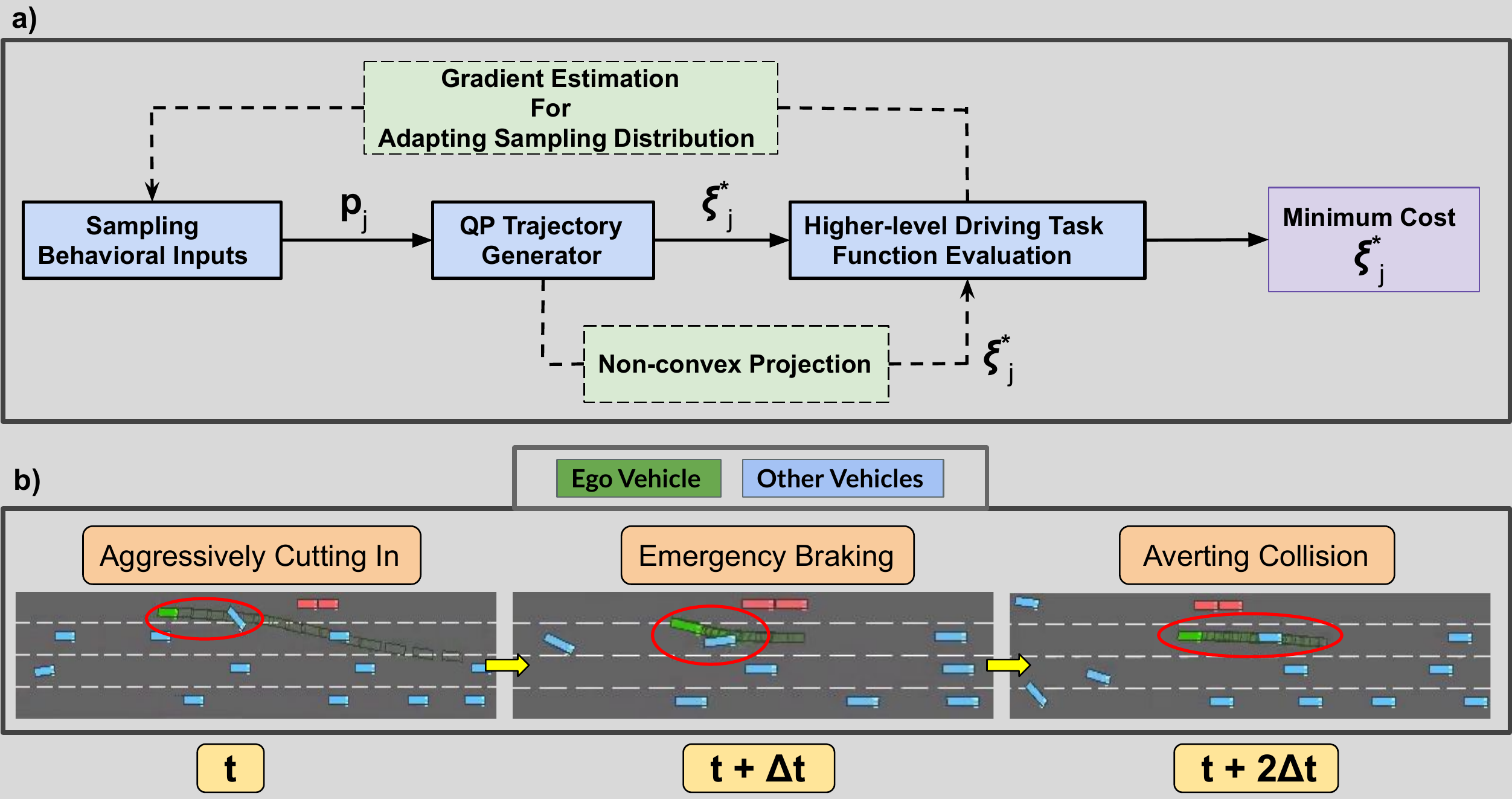}
    \caption{\textbf{Fig.(a)}: The solid blocks represent the components of the pipeline presented in works like \cite{li2021safe} \cite{wei2014behavioral}, while the blocks with dotted boundaries represent our contribution. Existing works draw behavioural inputs $\textbf{p}_j$ from a distribution and solve a simple QP trajectory planner for all those inputs. However, there is no mechanism to modify the behavioural input sampling based on the performance of the lower-level planner on the driving task. We address this issue by adding a gradient estimation block and a projection operator to aid constraint satisfaction. \textbf{Fig.(b)}   Our bi-level optimizer ensures safe driving in dense and potentially rash traffic. }
    \label{pipeline_fig}
    \vspace{-0.7cm}
\end{figure}

\begin{itemize}
    \item We propose a bi-level optimization where the upper-level variables represent the behavioural inputs while that at the lower level represent the associated motion plans.
    \item We combine Quadratic Programming (QP) with gradient-free optimization for solving the bi-level problem. Our approach avoids the common pitfalls of the Gradient descent-based approaches for bi-level optimization (see Section \ref{connections}). 
    \item We train a Conditional Variational Autoencoder with a differentiable optimization layer-based network architecture for warm-starting our bi-level optimizer.
    \item Our approach outperforms state-of-the-art Model Predictive Control (MPC) and RL-based approaches in safety metrics in dense traffic. 
\end{itemize}


\begin{figure*}[htp]
    \centering
    \includegraphics[scale = 0.28]{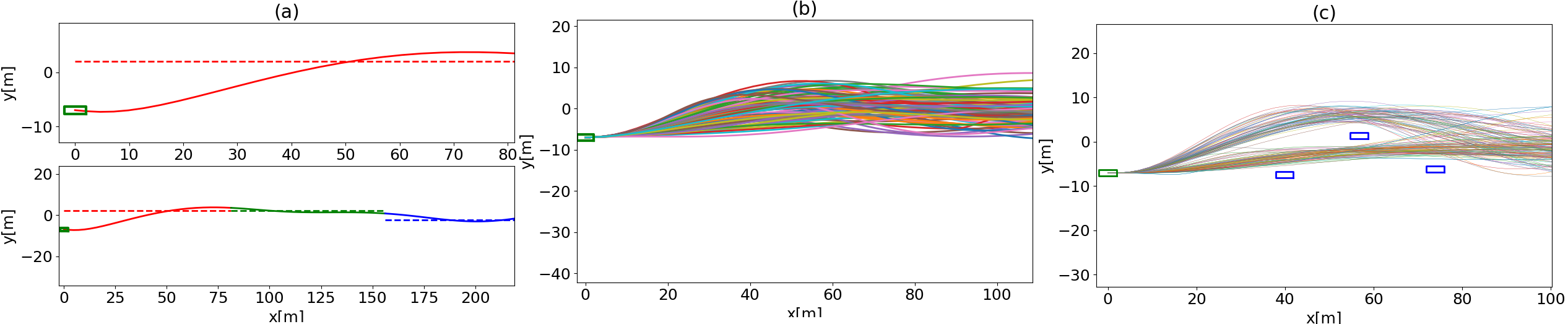}
    \caption{\textbf{Fig.(a),(b)}: A common parametrization for behavioural inputs is in the form of set-points for lateral offset (\textbf{Fig. (a) (top)}) \cite{hoel_rl_behavior}, \cite{rl_behavior_connected}. It can be used to induce lane-change manoeuvres in the ego vehicle. As shown in \textbf{Fig. (a)(bottom)}, for long horizon planning, we can split the trajectory segments into multiple parts and assign a lateral offset set-point to each of them. \textbf{Fig.(c)} shows different trajectories generated by sampling lateral offsets $y_{d, m}$ and forward velocity $v_{d, m}$ set-points from a Gaussian distribution and using them in \eqref{cost}-\eqref{ineq_const}. \textbf{Fig.(d)} showcases our custom batch optimizer that can solve the lower-level trajectory planning for all the sampled behavioural inputs in parallel. The figure represents the distribution of trajectories obtained in a cluttered environment, with a blue rectangle representing static parked vehicles. }
    \label{behavioural_description}
    \vspace{-0.5cm}
\end{figure*}

\section{Mathematical Preliminaries}
\subsubsection*{Symbols and Notation} Normal font lower-case letters will represent scalars, and bold font variants will represent vectors. The upper-case bold font letters will represent matrices. The superscript $T$ will denote the transpose of a matrix or a vector.


\subsection{Trajectory Parametrization}

\noindent Using the differential flatness of the bi-cycle car model, we aim to directly plan in the positional space ($x(t), y(t)$) of the ego-vehicle. Thus, we parametrize the position-level trajectory of the ego-vehicle in terms of polynomials in the following form:

\vspace{-0.3cm}

\small
\begin{align}
    \begin{bmatrix}
        x(t_1), \dots, x(t_m) 
    \end{bmatrix} = \textbf{W}\textbf{c}_{x},
     \begin{bmatrix}
        y(t_1), \dots, y(t_m) 
    \end{bmatrix} = \textbf{W}\textbf{c}_{y},
    \label{param}
\end{align}
\normalsize

\noindent where, $\textbf{W}$ is a matrix formed with time-dependent polynomial basis functions and $\textbf{c}_{x}, \textbf{c}_{y}$ are the coefficients of the polynomial. We can also express the derivatives in terms of $\dot{\textbf{W}}, \ddot{\textbf{W}}$. 


\subsection{Behavioral Input Parametrization}
\noindent We consider the following list of behavioural inputs in our formulation.

\begin{itemize}
    \item The planning horizon is split into $m$ parts and we assign a desired lateral offset set-point ($y_{d, m}$) to each of these segments (see Fig.\ref{behavioural_description}).Similarly, we assign longitudinal velocity set-points ($v_{d, m}$) to each of the $m$ segments
    
    
    \item Lastly, works like \cite{fernet_planner}, \cite{sun2022fiss} also include goal positions along longitudinal ($x_f$) and lateral directions ($y_f$) as behavioural inputs. 
\end{itemize}
\noindent We stack all the behavioural inputs into one parameter vector:
\vspace{-0.3cm}
\small
\begin{align}
\textbf{p} = \begin{bmatrix}
   y_{d, 1}, y_{d, 2},\dots, v_{d, 1}, v_{2, 2}, \dots v_{d, m}, x_f, y_f 
\end{bmatrix}.
\label{behaviour_param}
\end{align}
\normalsize

\noindent Note that not all elements of $\textbf{p}$ need to be used simultaneously in the lower-level trajectory planner. For example, \cite{hoel_rl_behavior}, \cite{rl_behavior_connected} use a single set-point for lateral offset and desired velocity as behavioural inputs. Our simulation results shown in Section \ref{sim} used four set-points for lateral offset and desired velocities to construct $\textbf{p}$. Nevertheless, \eqref{behaviour_param} represents all the behavioural parameterization that can be accommodated within our bi-level optimizer discussed in the next section.

\subsection{Lower-Level Trajectory Planning}

\noindent The lower-level trajectory optimization is formulated in the Frenet-frame: longitudinal ($x(t)$) and lateral ($y(t)$) motions of the ego-vehicle occur along and orthogonal to the reference centre line, respectively. 


\vspace{-0.3cm}
\small
\begin{subequations}
\begin{align}
  \min  \sum_t c_{s} (x(t), y(t)) +c_{l}(x(t), y(t))+c_v(x(t), y(t))\label{cost} \\
    (x(t_0), y(t_0), \dot{x}(t_0), \dot{y}(t_0), \ddot{x}(t_0), \ddot{y}(t_0) ) = \textbf{b}_0. \label{boundary_cond_initial}\\
    (x(t_f), y(t_f), \dot{y}(t_f)) = (x_f, y_f, 0). \label{final_boundary}\\
    g_j(x(t), y(t))\leq 0, \forall j, t \label{ineq_const} 
\end{align}
\end{subequations}
\normalsize
\vspace{-0.5cm}
\small
\begin{subequations}
\begin{align}
    c_{s} (x(t), y(t)) = (\ddot{x}(t))^2+\ddot{y}(t)^2\\
    c_{l}(x(t), y(t)) = \ddot{y}(t)-k_p(y(t)-\textbf{y}_d)-k_v\dot{y}(t))^2\\
    c_v(x(t), y(t)) = (\ddot{x}(t)-k_p(\dot{x}(t)-\textbf{v}_d)
\end{align}
\end{subequations}
\normalsize

\noindent The first term ($c_s(.)$) in the cost function \eqref{cost} ensures smoothness in the planned trajectory by penalizing high accelerations. The last two terms ($c_o(.), c_v(.)$) model the tracking of lateral offset ($y_{d, m}$) and forward velocity $(v_{d, m})$ set-points respectively and is inspired from works like \cite{hoel_rl_behavior}. For the former, we define a Proportional Derivative (PD) like tracking with gain $(k_p, k_v)$. For the velocity part, we only use a proportional term. The vector $\textbf{y}_d, \textbf{v}_d$ is formed by repeating each $y_{d, m}, v_{d, m}$ appropriate times and then vertically stacking them.  
Equality constraints \eqref{boundary_cond_initial} ensures that the planned trajectory satisfies the initial boundary conditions. Thus, vector $\textbf{b}_0$ is simply a stacking of initial position, velocity, and acceleration. The final boundary conditions are represented through constraints \eqref{final_boundary}. Inequalities \eqref{ineq_const} enforces collision avoidance, velocity, acceleration, centripetal acceleration, curvature bounds, and lane boundary constraints. We present the exact algebraic form of these constraints in Appendix \ref{gpu_batch_project}.



\section{Main Results}

\subsection{Proposed Bi-Level Optimization}
\noindent We formulate combined behavior and trajectory planning through the following bi-level optimization problem.
\small
\begin{subequations}
\begin{align}
    \min_{\textbf{p}} c_{u}(\boldsymbol{\xi}^{*}(\textbf{p}), \label{upper_cost}  \\
    \boldsymbol{\xi}^{*} \in \argmin_{\boldsymbol{\xi}} \frac{1}{2}\boldsymbol{\xi}^T\textbf{Q}\boldsymbol{\xi}+\textbf{q}^T(\textbf{p})\boldsymbol{\xi}_j, \label{lower_cost}  \\
    \textbf{A}_{eq}\boldsymbol{\xi} = \textbf{b}(\textbf{p}),  \qquad \textbf{g}(\boldsymbol{\xi}) \leq  \textbf{0}\label{lower_eq} 
\end{align}
\end{subequations}
\normalsize

\noindent where \eqref{lower_cost}-\ref{lower_eq} is the matrix representation of \eqref{cost}-\eqref{ineq_const} obtained using \eqref{param}. The behavioural inputs $\textbf{p}$ are defined in \eqref{behaviour_param}. Thus a part of it which comprises lateral offsets and forward velocity set-points enters the cost while the goal positions enter the affine equality constraints. The variable of the lower-level problem is $\boldsymbol{\xi} = (\textbf{c}_x, \textbf{c}_y)$.

As shown, we have an upper-level cost $c_u(.)$ that models the driving task. It depends on the optimal solution $\boldsymbol{\xi}^{*}$ computed from the lower-level trajectory optimization. The lower-level optimization explicitly depends on the parameter $\textbf{p}$ while the upper-level has an implicit dependency through $\boldsymbol{\xi}^{*}(\textbf{p})$. The goal of the lower-level optimizer is to compute an optimal solution for a given $\textbf{p}$. The upper level, in turn, aims to modify the parameter itself to drive down the upper-level cost associated with the optimal solution.

\subsection{Batch Optimization and Sampling-based Gradient for Bi-level Optimization}
\noindent Our main idea is to apply a gradient-free optimization technique on the upper-level cost. Alg.\ref{algo_1} summarizes the main steps of our proposed bi-level optimizer, wherein the left superscript $l$ is used to track variables across iterations. For example, ${^l}\boldsymbol{\mu}_p$ represents the mean of the sampling distribution at iteration $l$. On line 4, we draw $\overline{n}$ samples of $\textbf{p}_j$ from a Gaussian distribution with mean ${^{l}\boldsymbol{\mu}_p}$ and covariance ${^{l}\boldsymbol{\Sigma}_p}$. On line 6, we solve the lower-level trajectory optimization for each sampled parameter. Here, we adopt a two-step approach, a sample output of which is shown in Fig.\ref{behavioural_description}(c). In the first step, we solve the trajectory optimization without the inequality constraints, and then, in the second step, we project the obtained solution to the constrained set. On line 9, we compute the constraint residuals resulting from the optimal solutions. In line 10, we select the top $n$ samples with the least constraint residual to create $ConstraintEliteSet$. Line 11 constructs an augmented cost obtained by evaluating the upper-level cost $c_u(\boldsymbol{\xi}^*_j)$ on the samples from the $ConstraintEliteSet$ and adding the corresponding constraint residuals to it. On line 13, we select the top $q$ samples with the least augmented cost to construct the $EliteSet$. On line 14, we update the mean and variance of the sampling distribution of $\textbf{p}$ based on the samples of the $EliteSet$.



\subsubsection{Updating the Sampling Distribution} There are several ways to update the mean and variance on line 14 of Algorithm \ref{algo_1}. The simplest among these is to just fit a Gaussian distribution to the samples of $\boldsymbol{\xi}_j^*$ belonging to the $EliteSet$. However, this approach ignores the exact cost associated with the samples. Thus, in this work, we use the following update rule from sampling-based optimization proposed in \cite{bhardwaj2022storm}. 

\small
\begin{subequations}
\begin{align}
    {^{l+1}}\boldsymbol{\mu}_p = (1-\eta){^{l}}\boldsymbol{\mu}_p+\eta\frac{\sum_{j=1}^{j=q} s_j\textbf{p}_j   }{\sum_{j=1}^{j=q} s_j}, \label{mean_update}\\
    {^{l+1}}\boldsymbol{\Sigma}_p = (1-\eta){^{l}}\boldsymbol{\Sigma}_p+\eta\frac{ \sum_{j=1}^{j=q} s_j(\textbf{p}_j-{^{l+1}}\boldsymbol{\mu}_p)(\textbf{p}_j-{^{l+1}}\boldsymbol{\mu}_p)^T}   {\sum_{j=1}^{j=q} s_j} \label{cov_update}\\
    s_j = \exp{\frac{-1}{\gamma}(c_u(\boldsymbol{\xi}_j^*)+r_j(\boldsymbol{\xi}^{*}_j)} \label{s_formula}
\end{align}
\end{subequations}
\normalsize



\noindent where $\eta$ is the learning-rate and $\gamma$ is some scaling constant. As discussed in \cite{bhardwaj2022storm}, the update rules \eqref{mean_update} and \eqref{cov_update} are obtained by exponentiating the cost and then performing a sample estimate of its gradient with respect to the sampled argument (in this case $\textbf{p}$).

\vspace{-0.7cm}
\noindent 
 \begin{algorithm}[!h]
\caption{Bi-Level Optimization}
\small
\label{algo_1}
\SetAlgoLined
$N$ = Maximum number of iterations\\
Initiate mean $^{l}\boldsymbol{\mu}_{p}, ^{l}\boldsymbol{\Sigma}_{p}$, at $l=0$\\
\For{$l=1, l \leq N, l++$}
{
Draw $\overline{n}$ Samples $(\textbf{p}_1, \textbf{p}_2, \textbf{p}_j, ...., \textbf{p}_{\overline{n}})$ from $\mathcal{N}(^{l}\boldsymbol{\mu}_p, ^{l}\boldsymbol{\Sigma}_p)$\\
 \vspace{0.1cm}
Initialize $CostList$ = []\\
 \vspace{0.1cm}

Solve the lower-level trajectory optimization $\forall \textbf{p}_j$: \\
 \vspace{0.1cm}
\hskip1em \textbullet \hskip1em \text{Step 1: Solve the QP without inequalities}  
\vspace{-0.1cm}
\small
\begin{align*}
    \overline{\boldsymbol{{\xi}}}_j = \arg\min_{\boldsymbol{\xi}_j} \frac{1}{2}\boldsymbol{\xi}_j^T\textbf{Q}\boldsymbol{\xi}_j+\textbf{q}^T(\textbf{p}_j)\boldsymbol{\xi}_j\\
    \textbf{A}_{eq}\boldsymbol{\xi}_j = \textbf{b}(\textbf{p}_j)
\end{align*}
\normalsize
\hskip1em \textbullet \hskip1em \text{Step 2: Project to Constrained Set} \\
\vspace{-0.4cm}
\small
\begin{align*}
    \boldsymbol{{\xi}}^{*}_j = \arg\min_{\boldsymbol{\xi}_j} \frac{1}{2}\Vert \boldsymbol{\xi}_j-\overline{\boldsymbol{\xi}}_j\Vert_2^2\\
    \textbf{A}_{eq}\boldsymbol{\xi}_j = \textbf{b}(\textbf{p}_j), \qquad \textbf{g}(\boldsymbol{\xi}_j) \leq  \textbf{0}
\end{align*}
\normalsize

Define constraint residuals: $r_j(\boldsymbol{\xi}^{*}_j) = \sum\max(\textbf{0}, \textbf{g}(\boldsymbol{\xi}^{*}_j))$.\\
$ConstraintEliteSet  \gets$ Select top $n$   samples of $\textbf{p}_j, \boldsymbol{\xi}^*_j$ with lowest constraint residuals.\\
$cost \gets$ $c_u(\boldsymbol{\xi}_j^*)+r_j(\boldsymbol{\xi}^{*}_j)$,  over $ConstraintEliteSet$ \\
 \vspace{0.1cm}
append ${cost}$ to $CostList$ \\
 \vspace{0.1cm}
          
$EliteSet  \gets$ Select top $q$ samples of ($\textbf{p}_j, \boldsymbol{\xi}^*_j$)  with lowest cost from $CostList$.\\

$({^{l+1}}\boldsymbol{\mu}_p, {^{l+1}}\boldsymbol{\Sigma}_p ) \gets$ Update distribution based on $EliteSet$
 \vspace{0.1cm}
}
\Return{ Parameter $\textbf{p}_j$ and  $\boldsymbol{\xi}^{*}_j$ corresponding to lowest $c_u(\boldsymbol{\xi}^*_j)+r_j(\boldsymbol{\xi}^*_j)$ in the $EliteSet$}
\normalsize
\end{algorithm}

\vspace{-0.7cm}

\subsection{Computational Tractability of Alg.\ref{algo_1}}
\noindent The main computational bottleneck of Alg. \ref{algo_1} stems from the requirement of solving a large number ($\overline{n} \approx 1000$) of non-convex optimizations on line 6-8. However, each of these optimizations is decoupled from the other; thus, we can solve them in parallel (a.k.a., the batch setting) to ensure computational tractability. To this end, we first consider the QP presented on line 7. Solving it for the $j^{th}$ sample of $\textbf{p}_j$ reduces to following linear equations:

\small
\begin{align}
    \begin{bmatrix}
        \textbf{Q} & \textbf{A}^{T} \\ 
        \textbf{A} & \textbf{0}
    \end{bmatrix} \begin{bmatrix}
        \boldsymbol{\xi}_{j}\\ \boldsymbol{\mu}_{j}
    \end{bmatrix} = \begin{bmatrix}
        \textbf{q}(\textbf{p}_j)\\ \textbf{b}(\textbf{p}_j)
    \end{bmatrix}, \label{over_2}
\end{align}
\normalsize

\noindent where, $\boldsymbol{\mu}_j$ is the dual variable associated with the equality constraints. The left-hand side of \eqref{over_2} is constant and independent of the $\textbf{p}_j$. Thus, the solution for the entire batch can be constructed in one shot in the following manner:


\begin{figure}[!h]
    \centering
    \includegraphics[scale = 0.535]{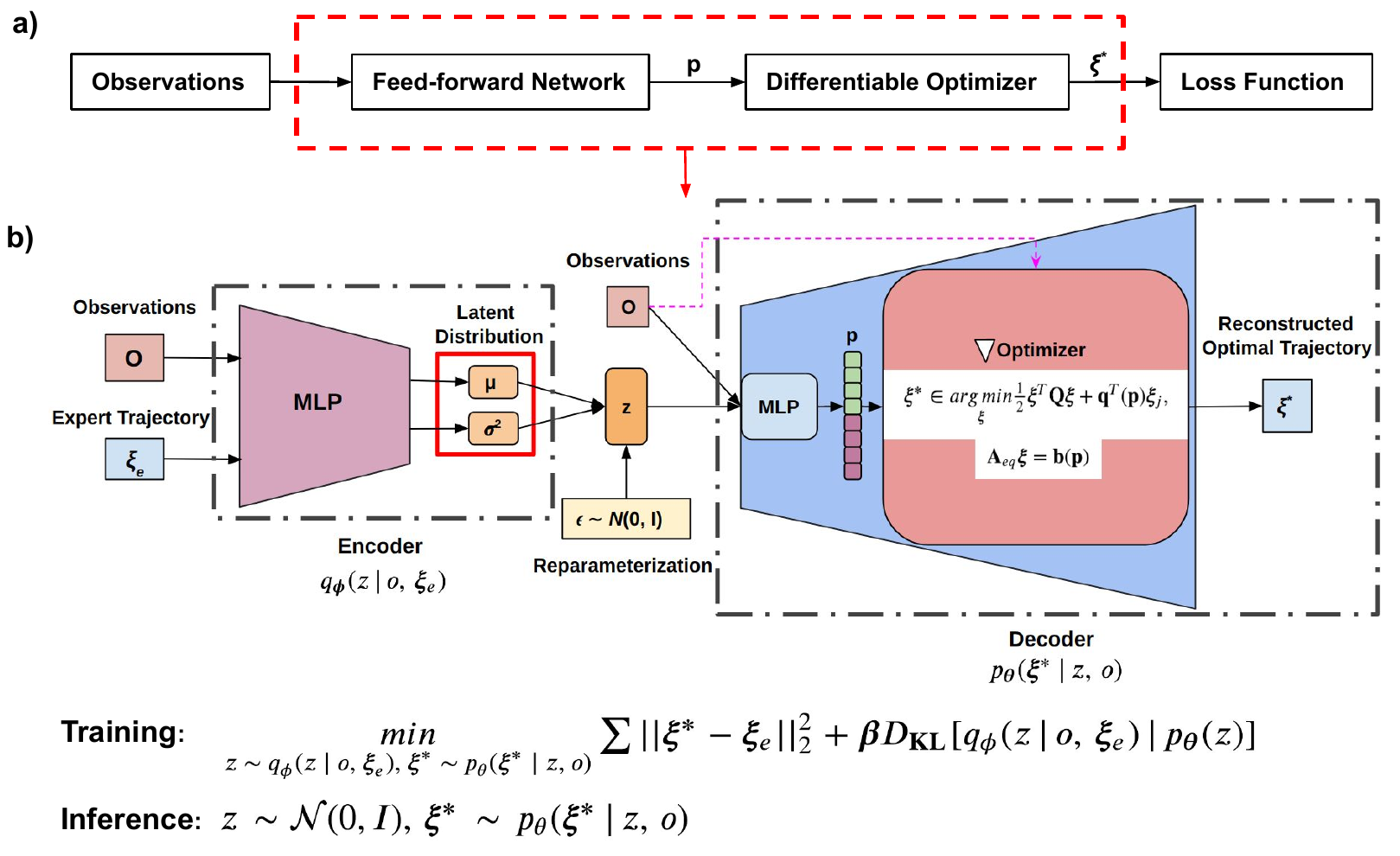}
    \caption{a) Our BC framework uses a combination of conventional feed-forward and differentiable optimization layers. b) To learn a good initialization distribution, a CVAE is trained to approximate the
    optimal distribution of $\textbf{p}$ that results in prediction $\boldsymbol{\xi}^*$ as close as possible to expert demonstrations $\boldsymbol{\xi}_e$.}
    \label{Overview_CVAE}
    \vspace{-1.3cm}
\end{figure}

\small
\begin{align}
\begin{bmatrix} 
\begin{array}{@{}c|c|cc@{}}
\boldsymbol{\xi}_{1} &...& \boldsymbol{\xi}_{\overline{n}} \\ \boldsymbol{\mu}_{1} &...& \boldsymbol{\mu}_{\overline{n}}
\end{array}
\end{bmatrix} =
    \overbrace{(\begin{bmatrix}
        \textbf{Q} & \textbf{A}^{T} \\ 
        \textbf{A} & \textbf{0}
    \end{bmatrix}^{-1}) }^{constant}\begin{bmatrix}
\begin{array}{@{}c|c|cc@{}}
    \textbf{q}(\textbf{p}_1) & ... & \textbf{q}(\textbf{p}_{\overline{n}})  \\
    \textbf{b}(\textbf{p}_1) & ... & \textbf{b}(\textbf{p}_{\overline{n}})
    \end{array}\end{bmatrix},
    \label{over_3}
\end{align}
\normalsize

\noindent The inverse on the right-hand side of \eqref{over_3} needs to be computed only once, irrespective of the batch size. Thus the batch solution of QP reduces to just a large matrix-vector product that can be trivially accelerated over GPUs.

For handling the projection operation on line 8 of Alg. \ref{algo_1}, we build on our recent work \cite{masnavi2022visibility}. This work showed that the core numerical algebra associated with projecting sampled trajectories on the collision avoidance constraints and motion bounds has the same batch QP structure as \eqref{over_2} and \eqref{over_3}. We extend \cite{masnavi2022visibility} to include curvature, centripetal acceleration bounds and lane boundary constraints while retaining its efficient batch projection update rule. We present a detailed derivation in Appendix \ref{gpu_batch_project}.


\vspace{-0.2cm}

\subsection{Learning Good Initialization Distribution} \label{nn_fitting}
\noindent In this section, we derive a  Behaviour Cloning (BC) framework to learn a neural-network policy that maps observations $\textbf{o}$ to optimal parameters $\textbf{p}$. Typically in BC, we assume that we have access to a data set $ (\textbf{o}, \boldsymbol{\xi}_e)$ that demonstrates the expert (optimal) trajectory $\boldsymbol{\xi}_e$ for each observation vector $\textbf{o}$. We cannot directly have access to a demonstration of the optimal behavioural parameter $\textbf{p}$ employed by the expert. Rather, we have their indirect observation through $\boldsymbol{\xi}_e$. Thus, our problem setting is more complicated than the typical BC setup. 

We address the mentioned challenges by using a network architecture that is a combination of conventional feed-forward, and differentiable optimization layers \cite{amos2017optnet}. An overview of the main concept is presented in Fig.\ref{Overview_CVAE} (a). The learnable weights are only present in the feed-forward layer. It takes in takes in observations $\textbf{o}$ to output the behavioural parameter $\textbf{p}$, which is then fed to an optimizer resulting in an optimal trajectory $\boldsymbol{\xi}^*$. The BC loss is computed over $\boldsymbol{\xi}^*$. The backpropagation required for updating the weights of the feed-forward layer needs to trace the gradient of the loss function through the optimization layer. 


\subsubsection*{Need for CVAE} We want our learned policy to induce a distribution over $\textbf{p}$ so that for each observation $\textbf{o}$, we can then draw samples from it to initialize our bi-level optimizer presented in Alg. \ref{algo_1} (line 4). With this motivation, we use a deep generative model called CVAE \cite{kingmaauto}, illustrated in Fig.\ref{Overview_CVAE} as our learning pipeline. It consists of an encoder and decoder architecture constructed from a multi-layer perceptron (MLP) with weights $\boldsymbol{\phi}$ and $\boldsymbol{\theta}$ respectively. Additionally, the decoder network has an optimization layer that takes the output ($\textbf{p}$) of its MLP to produce an estimate of an optimal trajectory $\boldsymbol{\xi}^*$. 


Let $\textbf{z}$ be a latent variable such that the $p_{\theta}(\textbf{z})$ represents a isotropic normal distribution($\mathcal{N}(0, \textbf{I})$). The decoder network maps this distribution to $p_{\theta}(\boldsymbol{\xi}^*|\textbf{z}, \textbf{o})$. The encoder network on the other hand maps $(\textbf{o}, \boldsymbol{\xi}_e)$ to a distribution $q_{\phi}(\textbf{z} | \textbf{o}, \boldsymbol{\xi}^{*})$ over $\textbf{z}$. In the offline phase, both the networks are trained end-to-end with loss function \eqref{eq_cvae}. The first term is the reconstruction loss responsible for bringing the output of the decoder network as close as possible to the expert trajectory. The second term in \eqref{eq_cvae} acts as a regularizer that aims to make the learned latent distribution $q_{\phi}(\textbf{z} | \textbf{o}, \boldsymbol{\xi}_e)$ as close as possible to the prior normal distribution. The $\boldsymbol{\beta}$ hyperparameter acts as a trade-off between the two cost terms. The detailed architecture of our CVAE, alongside the training hyperparameters, is presented in Appendix. \ref{CVAE Hyperparams}.

\vspace{-0.3cm}
\small
\begin{equation}
    {\textit{L}}_{\textit{CVAE}} = \underset{\boldsymbol{\theta}, \boldsymbol{\phi}}{min}\sum \Vert \boldsymbol{\xi}^*(\boldsymbol{\theta}, \boldsymbol{\phi}) - \boldsymbol{\xi}_e \Vert_2^2 \newline
    + \beta \, {D}_{\mathbf{KL}}[q_\phi(\textbf{z}\,| \,\textbf{o}, \,\boldsymbol{\xi}_e)\, | \,p_{\theta}(\textbf{z} )]  \label{eq_cvae}
\end{equation}
\normalsize
\vspace{-0.5cm}

In the inferencing (online) phase, we draw samples of $\textbf{z}$ from an isotropic normal distribution and then pass them through the decoder MLP to get samples of optimal behavioural parameter $\textbf{p}$. Finally, the parameter samples are passed through the optimizer to generate distribution for the optimal trajectory $\boldsymbol{\xi}^*$.



\begin{figure*}[htp]
    \centering
    \includegraphics[scale = 0.385]{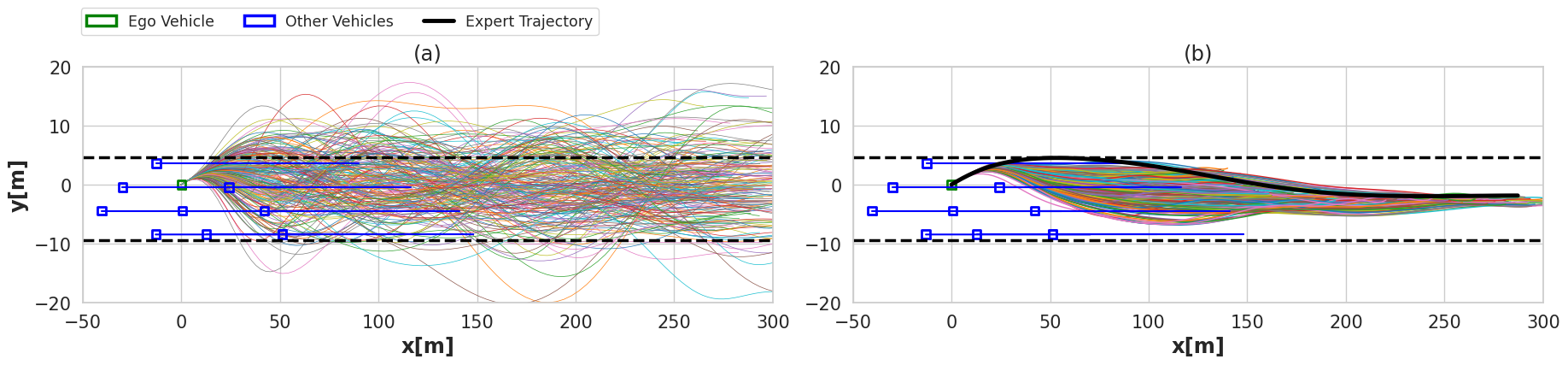}
    \caption{Fig.(a) shows the trajectory distribution resulting from sampling behavioural inputs $\textbf{p}$ from a Gaussian distribution. Fig.(b) shows the corresponding distribution when sampling $\textbf{p}$ from a learned CVAE. As can be seen, CVAE results in a more structured and smoother trajectory distribution concentrated around the expert demonstration, shown in black. The trajectory samples also conform to lane boundaries shown in dotted black lines. The blue rectangle represents the neighbouring vehicles moving along straight-line (blue) trajectories. The green rectangle represents the ego-vehicle. }
    \vspace{-0.2cm}
    \label{learned_sampler}
    \vspace{-0.5cm}
\end{figure*}

\subsubsection{Choice of Differentiable Optimizer} Ideally, we should embed the entire lower-level trajectory optimizer \eqref{lower_cost}-\eqref{lower_eq} into the CVAE decoder architecture. However, backpropagating through such non-convex optimization is fraught with technical difficulties. In fact, the greatest success of learning with optimization layers has come while embedding convex optimizers into neural networks \cite{amos2017optnet}. Thus, we adopt a simplification as shown in Fig.\ref{Overview_CVAE}(b). We construct our optimiztion layer with convex cost \eqref{lower_cost} and affine equality constraints in \eqref{lower_eq}, both of which depend explicitly on $\textbf{p}$. We ignore the inequality constraints, where the parameters do not appear. The intuition behind our choice is that the expert trajectory will be collision-free and kinematically feasible and thus would automatically satisfy the inequality constraints. Thus, we need to figure out the right set of $\textbf{p}$ that can mimic the expert behaviour as closely as possible. 

\subsubsection{Example of a  CVAE Output} Fig.\ref{learned_sampler} contrasts initialization of Alg. \ref{algo_1} from a naive Gaussian distribution and our learned CVAE. For Fig.\ref{learned_sampler}(a), we sampled $\textbf{p}$ from a Gaussian distribution and solved the QP presented in line 7 of Algorithm \ref{algo_1}, resulting in a distribution of trajectories. We repeated the same process for the samples drawn from CVAE in Fig.\ref{learned_sampler}(b). It can be clearly seen that the distribution resulting from  $\textbf{p}$ drawn from CVAE is smoother, conforms to lane boundaries, and is concentrated around the expert demonstration trajectory.


\section{Connections to Existing Works} \label{connections}
\subsubsection*{Trajectory Sampling Approaches} Existing works like \cite{wei2014behavioral}, \cite{fernet_planner} can be viewed as a special case of our Alg. \ref{algo_1} obtained by performing only one iteration of the bi-level optimizer. These cited works sample the parameter $\textbf{p}$ (lateral offsets, forward velocity, etc.), albeit not from a Gaussian distribution but a pre-discretized grid. This is followed by the execution of lines 6-8 and ranking of the upper-level cost (line 13) associated with the generated trajectories. However, \cite{wei2014behavioral}, \cite{fernet_planner} do not have any mechanism to adapt the sampling distribution (or grid) to reduce the upper-level cost. Authors in \cite{sun2022fiss} address this drawback to some extent as they adapt the sampling strategy based on optimal trajectories obtained in the past planning cycles. Such an adaptation strategy would be akin to performing one iteration of Alg. \ref{algo_1} and then warm-starting the sampling distribution of $\textbf{p}$ at the next planning cycle with the updated mean and variance obtained from line 14. 

The lower-level planners of \cite{wei2014behavioral}, \cite{fernet_planner}, \cite{sun2022fiss} ignore inequality constraints and essentially solve the QP presented on line 7 of Alg. \ref{algo_1}. The constraint residuals (e.g. obstacle clearance) are augmented into the cost function, similar to line 11. Our current work includes an additional projection operator at line 8 of Algorithm \ref{algo_1} to aid in constraint satisfaction.


\subsubsection*{RL Based Approaches} Works like \cite{hoel_rl_behavior}, \cite{rl_behavior_connected}, \cite{huegle2019dynamic}, \cite{li2021safe} can be viewed as training a function approximator to learn the solution space of the bi-level optimizer presented in \ref{algo_1}. In Section \ref{nn_fitting}, we have made a similar attempt using a supervised setting. The RL approaches of \cite{hoel_rl_behavior}, \cite{rl_behavior_connected}, \cite{huegle2019dynamic}, \cite{li2021safe} would achieve this in a self-supervised setting based on just feedback of reward (upper-level cost) from the environment.  

\subsubsection*{ Bi-level Optimization} The bi-level approaches are extensively used in motion planning. For example, see \cite{sun2020fast}, \cite{song2022policy}. Our work is closely related to the latter. In \cite{song2022policy}, an offline bi-level optimization is used to generate optimal higher-level behaviours (parameter $\textbf{p}$) for drones and subsequently, a neural network is trained to learn this solution space. At run-time, the neural network's output is the true solution. In sharp contrast, we take the output from the CVAE trained in Section \ref{nn_fitting} as just a guess for the optimal parameter and adapt it in real-time in Alg.\ref{algo_1}. 

\subsubsection*{Comparison with Gradient Descent} Bi-level optimizations are commonly solved through Gradient Descent \cite{sun2020fast}. It requires computing the Jacobian of the optimal solution $\boldsymbol{\xi}^*$ with respect to parameter $\textbf{p}$ through implicit differentiation \cite{amos2017optnet}. The main drawback of this approach is that implicit differentiation has technical difficulties in case there are multiple local minima and/or when the lower-level problem is infeasible. In contrast, our Alg. \ref{algo_1}  does not require the lower-level optimization to be feasible and allows for its early termination. In either case, the constraint residuals can measure the quality of the optimal trajectory, which is why we augment it into the upper-level cost on line 11.

\section{Experiments}\label{sim}



\subsection{Implementation Details}
\noindent We implemented Alg. \ref{algo_1} including the lower-level optimizer in Python using JAX \cite{jax} library as our GPU-accelerated linear algebra back-end. The matrix $\textbf{W}$ in \eqref{param} is constructed from a $10^{th}$ order polynomial. Our simulation pipeline was built on top of the Highway Environment (highway-env) simulator \cite{Leurent_An_Environment_for_2018}. The neighbouring vehicles used IDM \cite{idm_paper} for longitudinal and MOBIL \cite{kesting2007general} for lateral control.

\subsubsection{Hyper-parameter Selection} The sampling size $\overline{n}$ in Alg. \ref{algo_1} was 1000. The $ConstraintEliteSet$ (line 10, Alg. \ref{algo_1}) and $EliteSet$ (line 13, Alg. \ref{algo_1}) had 150 and  50 samples respectively. For our bi-level optimizer, the behavioural input $\textbf{p}$ was modelled as four set-points for lateral offsets and desired longitudinal velocities. That is, $\textbf{p} = \begin{bmatrix} y_{d, 1}, \dots, y_{d,4}, v_{d, 1}, \dots, v_{d,4}   \end{bmatrix}$. We divided the planning horizon into four segments and associated one pair of lateral offset and desired velocity to each of these. We recall that the Alg. \ref{algo_1} computes the optimal $\textbf{p}$ along with the associated trajectory. $\gamma$ value in \eqref{s_formula} was 0.9.



\begin{figure}[htp]
    \centering
    \includegraphics[scale = 0.34]{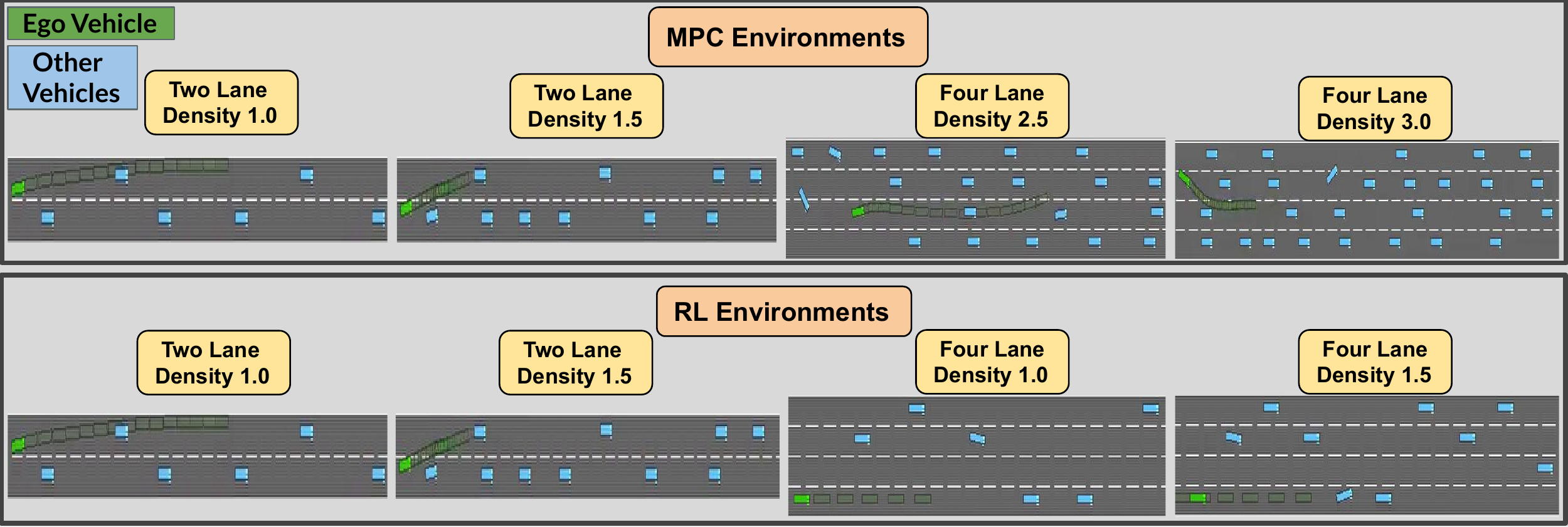}
    \caption{Two-lane and four-lane driving scenarios with varying traffic density for benchmarking our approach with MPC and RL baselines.}
    \label{env_bench}
    \vspace{-0.3cm}
\end{figure}

\subsubsection{CVAE Training} The details of the encoder and decoder network architecture of our CVAE are presented in the accompanying video and Appendix \ref{cvae_appendix}. During training, the input to the CVAE is the expert trajectory and a 55-dimensional observation vector ($\textbf{o}$) containing the state of the ego-vehicle, the ten closest obstacles and the road boundary. For the ego-vehicle, the state consists of a heading, lateral and longitudinal speeds. The state consists of longitudinal and lateral positions and the corresponding velocities for the ten closest obstacles. We express all the position-level information with respect to the centre of the ego vehicle. During inferencing, the decoder network only needs $\textbf{o}$ and samples of $\textbf{z}$ drawn from an isotropic Gaussian.

We used the cross-entropy method, run offline with a batch size of 1000, to collect the demonstration of optimal trajectories for training our CVAE. We note that our demonstrations could possibly be sub-optimal. However, even with such a simple data set, our CVAE is able to learn useful initialization for our bi-level optimizer presented in Alg.\ref{algo_1}. 






\subsubsection{MPC Baselines} We also used Alg.\ref{algo_1} in a receding horizon manner to create an MPC variant of our bi-level optimizer. We will henceforth refer to it as MPC-Bi-Level. It takes the same observation vector $\textbf{o}$ as the CVAE and outputs polynomial coefficients of the optimal trajectories. These are converted to steering and acceleration input vectors, and the ego-vehicle executes the first five elements of these in the open loop before initiating the next re-planning. We compare our MPC-Bi-Level with the following baselines.

\begin{itemize}
    \item MPC-Vanilla: This baseline runs without a behavioural layer. We only have the lower-level optimization  \eqref{lower_cost}-\eqref{lower_eq}. The parameter $\textbf{p}$ is a scalar representing the set-point for the desired longitudinal velocity.
    \item MPC-Grid: This baseline uses the same set of behavioural parameters $\textbf{p}$ as our MPC-Bi-Level but samples them from a pre-specified fixed grid.
    \item MPC-Random: This baseline is similar to MPC-Grid but samples $\textbf{p}$ from a Gaussian distribution.
    \item Batch-MPC of \cite{adajania2022multi}: This baseline is similar to MPC-Grid but uses a different set of behavioural inputs, namely goal positions for the longitudinal and lateral components of the trajectory. That is, $\textbf{p} = \begin{bmatrix} x_f, y_f  \end{bmatrix}$. 
\end{itemize}


\begin{figure*}[htp]
    \centering
    \includegraphics[scale = 0.28]{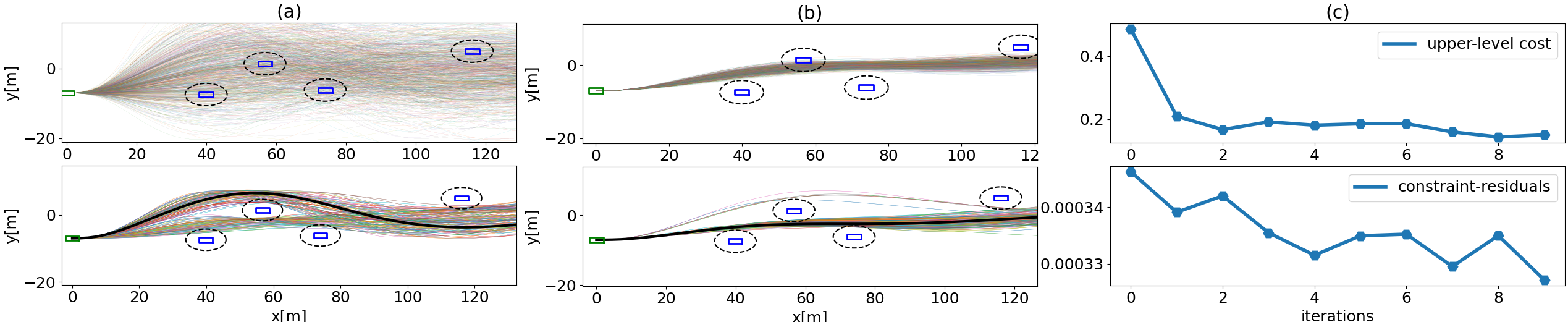}
    \caption{Empirical validation of Alg.\ref{algo_1}. The top of Fig.(a),(b) shows the trajectory distribution resulting from sampling behavioural inputs $\textbf{p}$ from a Gaussian distribution and solving the QP at line 7 of Alg.\ref{algo_1}. The bottom plots in both figures show the modified distribution after projection onto the constrained set. Fig.(a),(b) correspond to the distribution at the first and the fifth iteration, respectively. As the iterations progress, the variance of the trajectory distribution shrinks, and the upper-level cost (Fig.(c), top) saturates, indicating convergence of our bi-level optimizer. Fig.(c) (bottom) shows the constraint residual of the lower-level optimization at each iteration. }
    \label{qual_result_basic}
    \vspace{-0.3cm}
\end{figure*}

\noindent \subsubsection{RL Baselines } We also compare our approach against Deep Q-Network (DQN) and Proximal Policy Optimization (PPO), developed using the framework outlined in  \cite{hoel2018automated} and \cite{liu2020decision} respectively. 
The input observation is the same as our MPC-Bi-Level. The action space is discrete with 5 different behaviours, namely \textit{faster}, \textit{slower}, \textit{left-lane change}, \textit{idle}, \textit{right-lane change}. These behaviours are then mapped to a set-point for lateral offset or longitudinal velocity and tracked through a PID controller using appropriate steering and acceleration commands.
Both DQN and PPO have been trained on the highway-env simulator using Stable-Baselines 3 \cite{stable-baselines3}. See Appendix \ref{rl_appendix} for further details.

\subsubsection{Environments, Tasks, and Metrics}
\noindent The driving scenarios are presented in Fig.\ref{env_bench}. For each scenario, we had two different traffic densities. We use the internal parameter of HighEnv named "density" to control how closely each vehicle is placed at the start of the simulation. The RL baselines did not perform well in very dense environments and thus were tested in sparser environments than the MPC-based approaches. In each scenario, we evaluated 50 episodes with different randomly initialized traffic. We fixed the random seed of the simulator to ensure that all RL and MPC baselines are tested across the same set of traffic configurations.

The task in the experiment was for the ego-vehicle to drive as fast as possible without colliding with the obstacles and going outside the lane boundary. Thus, the upper-level cost of our bi-level optimizer has the form ${(\sqrt{\dot{x}^{*^2(t)}+\dot{y}^{*^2(t)}}}-v_{max})^2$, where $({x}^{*(t)}, {y}^{*(t)}) $ are the optimal velocity profiles obtained from the lower-level optimization. Note that these are obtained from $\boldsymbol{\xi}^*$ through relationship \eqref{param}. The safety is handled by the constraints of the lower-level optimization. Our evaluation metric has two components: (i) collision rate and (ii) average velocity achieved within an episode. Since the ego-vehicle can achieve arbitrary high velocity while driving rashly, we only consider velocities from collision-free episodes.



\subsection{Empirical Validation of Convergence}
\noindent Fig.\ref{qual_result_basic} shows the performance of our bi-level optimizer presented in Alg.\ref{algo_1} on a typical environment with static obstacles. The top plot of Fig.(a), (b) shows the trajectory distributions at the first and the fifth iteration, respectively, resulting from sampling the behavioural inputs $\textbf{p}$ from a Gaussian distribution and solving the QP presented in line 7 of Alg.\ref{algo_1}. The bottom plots in these figures show how the distribution changes when we project them onto the feasible set of collision avoidance and kinematic constraints. The following key observations can be made from Fig.\ref{qual_result_basic}:

\begin{itemize}
    \item The projection operation results in trajectories residing in different homotopies, which in turn proves crucial for proper exploration of the state space.
    
   \item The variance of the trajectory distribution shrinks, and the upper-level cost (Fig.\ref{qual_result_basic}(c)(top)) saturates. This is a typical convergent behaviour observed in sampling-based optimizers such as Alg.\ref{algo_1}. Please note that shrinking of trajectory variance in Fig.\ref{qual_result_basic}(a), (b) (top) also signifies that the sampling distribution for $\textbf{p}$ has also converged to an optimal one.
    
    \item Finally,  Fig.\ref{qual_result_basic}(c)(bottom) validates the role of our projection operator.
\end{itemize}


\subsection{Importance of Bi-Level Adaptation} \label{mpc_benchmark}

\begin{figure}[htp]
    \centering
    \includegraphics[scale = 0.36]{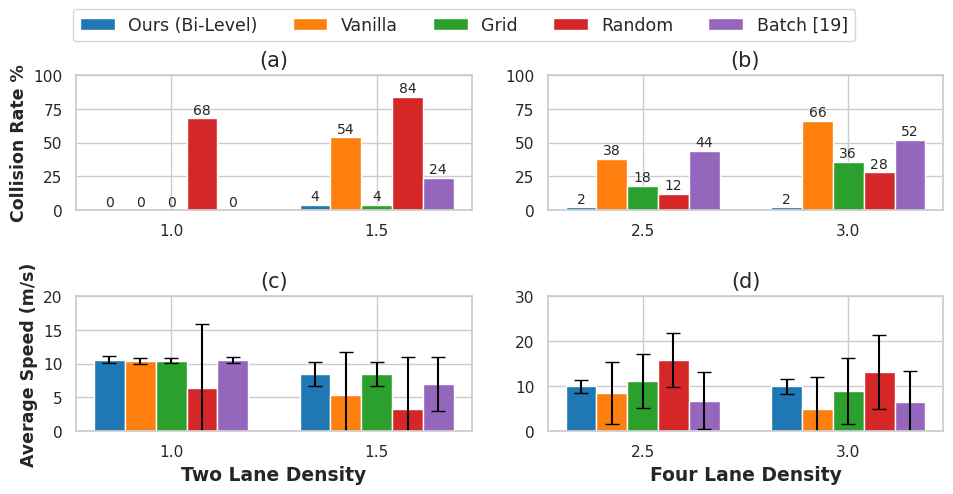}
    \caption{Comparison of MPC Bi-Level (Ours) with other MPC-based baselines in two (a, c) and four-lane (b, d) driving scenarios.}
    \label{mpc_comparison}
    \vspace{-0.3cm}
\end{figure}

\noindent In a two-lane driving scenario, there is only a limited set of manoeuvres that the ego-vehicle can do. Thus, on a low traffic density, all the baselines except MPC-Random achieve perfect collision rate (Fig.\ref{mpc_comparison} (a)). This shows that it is not critical to have a dedicated behavioural layer on simple driving scenarios. Our observation is not surprising as existing results like \cite{gutjahr2016lateral}, \cite{lin2022model} have shown promising results without explicitly incorporating behavioural inputs, but in very sparse environments. Fig.\ref{mpc_comparison}(a) also shows that in a sparse two-lane environment, a simple grid-search used by MPC-Grid and Batch-MPC \cite{adajania2022multi} is enough to come up with the right set of behavioural inputs. However, as traffic density increases in the two-lane setting, the safety improvement provided by our MPC-Bi-Level becomes distinctly apparent. The trend is particularly stark in highly dense four-lane driving scenarios, where our approach provides a 4-10x reduction in collision rate over other baselines. Fig.\ref{mpc_comparison}(b), (d) shows that the average speed achieved by our MPC-Bi-Level is generally either better or competitive with all the baselines.

\subsection{Safety Improvements over RL} \label{rl_benchmark}
\begin{figure}[htp]
    \centering
    \includegraphics[scale = 0.36]{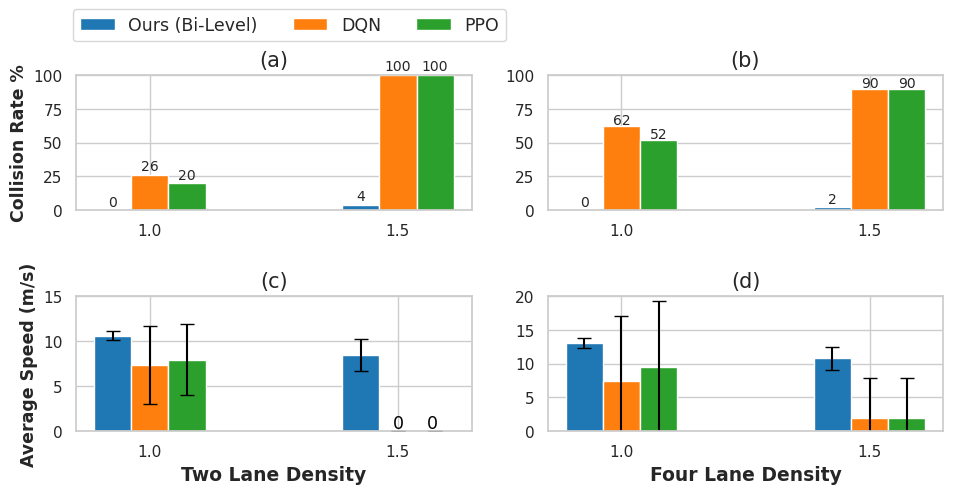}
    \caption{Comparison of MPC Bi-Level (Ours) with other RL-based baselines in two (a, c) and four-lane (b, d) driving scenarios.}
    \label{rl_compare}
    \vspace{-0.3cm}
\end{figure}

\noindent We trained both DQN and PPO for over 5 million steps. However, we could make them work reasonably in only sparse two-lane traffic. Nevertheless, the collision rate and velocity (Fig.\ref{rl_compare}(a, c) ) achieved by DQN and PPO were still drastically worse than our MPC-Bi-Level. The performance gap increased further in a dense four-lane setting, as shown in Fig.\ref{rl_compare}(b, d).

\subsection{Effect of CVAE Initialization}

\begin{figure}[htp]
    \centering
    \includegraphics[scale = 0.36]{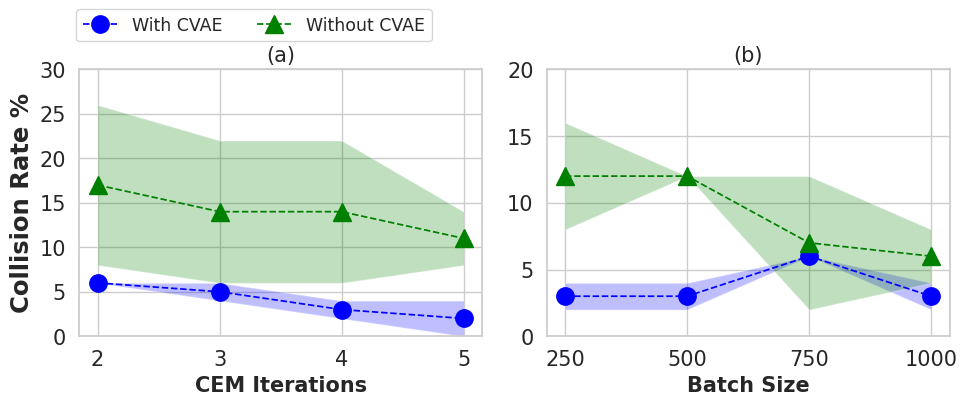}
    \caption{Computational advantage achieved by initializing Alg. \ref{algo_1} through samples drawn from learnt CVAE over baseline Gaussian distribution. CVAE allows us to achieve a better collision rate with a smaller batch size. } 
    \label{cvae_effect}
    \vspace{-0.4cm}
\end{figure}

\noindent Fig.\ref{cvae_effect}(a) shows that for a relatively small batch size of 250, the learned CVAE initializer achieves a 4x reduction in collision rate over the naive Gaussian distribution. However,  the performance gap between both initializations reduces a bit as we increase the batch size (Fig.\ref{cvae_effect}(b)). Thus, both Fig.\ref{cvae_effect}(a)-(b) validate that a learned CVAE initializer can be particularly helpful when Alg. \ref{algo_1} is run with a limited computation budget or on resource-constrained hardware, wherein we have to contend with a smaller batch size for Alg.\ref{algo_1}.

\vspace{-0.4cm}

\subsection{Computation Time}
\begin{figure}[htp]
    \centering
    \includegraphics[scale = 0.365]{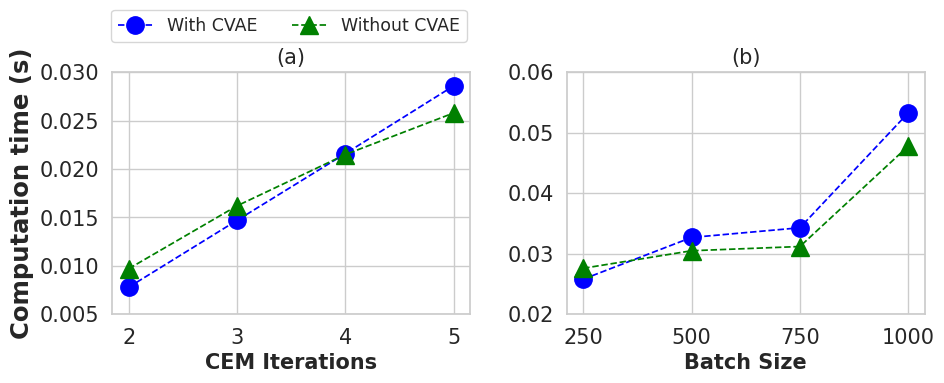}
    \caption{Samples drawn from learnt CVAE distribution leads to better collision-rate without the expense of computational time. } 
    \label{comp_time}
    \vspace{-0.4cm}
\end{figure}

\noindent Fig.\ref{comp_time} shows the computation time requirement for our Alg.\ref{algo_1} on a laptop with RTX 3080 GPU. In sparse traffic scenarios, two iterations of Alg.\ref{algo_1} proved enough to achieve a low collision rate when initialized with the learned CVAE. For a batch size of 250, this corresponds to a feedback rate of around 100 Hz. All the benchmarking presented in Section \ref{mpc_benchmark}, \ref{rl_benchmark} were obtained with the same batch size but used 5 iterations of Alg.\ref{algo_1}, totalling to $0.03s$. Fig.\ref{comp_time}(b) demonstrates a moderate increase in the computation time with respect to batch size; even for a batch size of 1000, the computation time was less than $0.06s$. Finally, both Fig.\ref{comp_time}(a), (b) shows that the additional overhead of inferencing the learned CVAE is very minimal.



\section{Conclusion and Future Work} 
We proposed a novel bi-level optimization that can simultaneously search for the optimal higher-level behavioural decisions along with the lower-level trajectories necessary for executing them. Our custom optimizer combines features from gradient-free, sampling-based optimization with QP and runs in real time due to an efficient GPU parallelization of the lower-level optimization. We also proposed a CVAE architecture constructed from a feed-forward neural network and differentiable optimization layers to learn good initialization for our bi-level optimizer. 

We conducted extensive experiments to showcase the importance of having a dedicated behavioural layer. Our approach also outperformed
competing MPC and RL baselines. Finally, the learned CVAE initialization improved the computational tractability of our bi-level optimizer by reducing the batch size and number of iterations required to achieve a given collision rate.

Our bi-level optimizer sets the groundwork for an RL framework where the policy is a combination of a neural network that provides higher-level decisions and a local MPC. We conjecture that an Augmented Random Search technique \cite{mania2018simple} for training both the network and MPC in an end-to-end fashion will have a very similar structure as Alg.\ref{algo_1}. 


\bibliography{main}
\bibliographystyle{IEEEtran}

\appendices
\section{A. CVAE Architecture \& Training Hyperparameters} \label{cvae_appendix}
\label{CVAE Hyperparams}

\begin{table}[!h]
\begin{subtable} 
\centering
\begin{tabular}{|c|c|c|c|}
\hline
\multicolumn{4}{|c|}{\textbf{Encoder Network}} \\ \hline
Block & \vtop{\hbox{\strut Layers}} & \vtop{\hbox{\strut Output Size}} & \vtop{\hbox{\strut Activation}} \\ \hline
MLP 1 - 4 & Linear, Batchnorm & 1024 & ReLU  \\ \hline
MLP 5 & Linear, Batchnorm & 256 & ReLU  \\ \hline
Mean & Linear & 2  & None  \\ \hline
Variance & Linear & 2 & Softplus  \\ \hline

\end{tabular}
\vspace{0.25cm}
\end{subtable}

\begin{subtable}
\centering
\begin{tabular}{|c|c|c|c|}
\hline
\multicolumn{4}{|c|}{\textbf{Decoder Network}} \\ \hline
Block & \vtop{\hbox{\strut Layers}} & \vtop{\hbox{\strut Output Size}} & \vtop{\hbox{\strut Activation}} \\ \hline
MLP 1 - 4 & Linear, Batchnorm & 1024 & ReLU  \\ \hline
MLP 5 & Linear, Batchnorm & 256 & ReLU  \\ \hline
\textbf{p} & Linear & 8  & None  \\ \hline
$\boldsymbol{\xi^*}$ & Optimization Layer & 22  & None  \\ \hline

\end{tabular}
\end{subtable}
\caption{CVAE architecture composed of Encoder and Decoder Networks illustrated in Fig.\ref{Overview_CVAE}}
\vspace{-0.3cm}
\end{table}

The optimizer used for training the CVAE was AdamW \cite{loshchilov2018decoupled} with a learning rate of 1e-4 and weight decay of 6e-5 for a total of 80 epochs. Moreover, the learning rate was decayed by $\gamma=0.1$ every 10 epochs. We tackle the KL-vanishing issue by applying a monotonic annealing schedule of $\boldsymbol{\beta}$ coefficient \cite{bowman-etal-2016-generating}, starting with 0 and gradually annealing the $\boldsymbol{\beta}$ at each step.

\section{B. Additional Experiments on Two-Way Environment}
\label{Two Way}

\begin{figure}[htp]
    \centering
    \includegraphics[scale = 0.34]{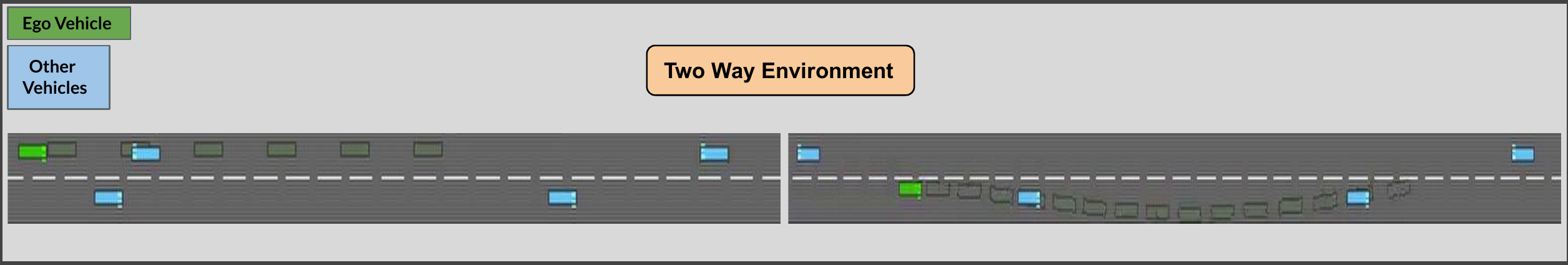}
    \caption{Two-way highway driving scenario for benchmarking our approach with MPC and RL-based baselines.}
    \label{Two-Way_Env}
    \vspace{-0.3cm}
\end{figure}

\begin{figure}[htp]
    \centering
    \includegraphics[scale = 0.375]{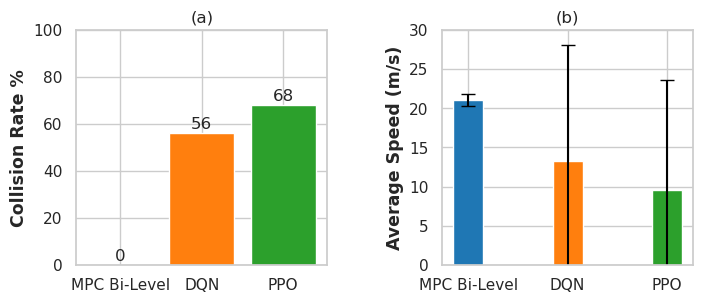}
    \caption{Comparison of MPC Bi-Level with other RL-based baselines in two-way (a, b) driving scenario.}
    \label{Two-way_Comp}
    \vspace{-0.3cm}
\end{figure}

DQN and PPO fail to work reasonably in a relatively simple Two-Way highway driving scenario Fig.\ref{Two-Way_Env}. MPC-based approaches, including our MPC Bi-Level, achieved zero collision rate and higher velocity than RL baselines, as shown in Fig.\ref{Two-way_Comp} (a, b).

\section{RL Training Hyperparameters} \label{rl_appendix}
The policy estimators employed in RL baselines are MLPs with two hidden layers of 256 neurons each. The action space is discrete with the following different behaviours:

\begin{itemize}
    \item \textit{faster}: Increase velocity by $5$ m/s
    \item \textit{slower}: Decrease velocity by $5$ m/s
    \item \textit{left-lane change}: change lateral position by $4$ m to the left of the current position.
    \item \textit{idle}: Keep moving with the current velocity and lateral offset set-points.
    \item \textit{right-lane change}: change lateral position by $4$ m to the right of the current position.
\end{itemize}
The complete set of RL training hyper-parameters are summed up in Table \ref{tab:train_param}.

\begin{table}
\begin{center}
\caption{RL baselines training parameters}
\label{tab:train_param}
\setlength{\tabcolsep}{3pt}
\begin{tabular}{ c c c }
\hline
Agent & Parameter& Value \\
\hline
& Number of training steps & 5M\\
& Policy Scheduling Time & 1s \\
& Input neurons & 55 \\
& Hidden layers  & 2\\
& Hidden layers neurons & 256 \\
& Output neurons & 5 \\
& Discount factor & 0.8 \\
& Learning rate & 5e-4\\
\hline
DQN & Replay Memory size & 15k \\
& Initial exploration constant  &  1\\
& Final exploration constant & 0.1  \\
& Target Network update frequency & 50\\
 & Batch size & 32 \\
 \hline
 PPO & number of steps & 10 \\
 & Batch size & 64 \\
 & Generalized Advantage Estimation(GAE) $\lambda$ & 0.95 \\
 & clipping coefficient & 0.2 \\
 & value-function coefficient & 0.5 \\
\hline
\end{tabular}
\label{tab:rl_hyper}
\end{center}
\end{table}

\section{GPU Accelerated Batch Projection} \label{gpu_batch_project}
In this section, we modify the projection operator of \cite{masnavi2022visibility} to make it suitable for car-like vehicles while retaining the batch QP structure presented in \eqref{over_3}. We begin by summarizing the set of inequality constraints used in the lower-level optimization ($g_j(x(t))\leq 0 $) in Table \ref{ineq_list}. 

\begin{table*}[!t]
\centering
\caption{\scriptsize{List of Inequality Constraints Used in the lower level optimization}}
\small
\begin{tabular}{|c|c|c|c|c|c|}
\hline
Constraint Type & Expression & Parameters   \\ \hline
Collision Avoidance  & $-\frac{(x(t)-x_{o, i})^2}{a^2}-\frac{(y(t)-y_{o, i})^2}{b^2}+1\leq 0$ & \makecell{$\frac{a}{2}, \frac{b}{2}$: axis of the circumscribing ellipse of vehicle footprint. \\ $x_{o,i}(t), y_{o, i}(t)$: trajectory of neighboring vehicles} \\ \hline
Velocity bounds & $\sqrt{\dot{x}(t)^2+\dot{y}(t)^2}\leq v_{max}$ & $v_{max}$: maximum velocity of the ego-vehicle   \\ \hline
Acceleration bounds & $\sqrt{\ddot{x}(t)^2+\ddot{y}(t)^2}\leq a_{max}$ & $a_{max}$: maximum acceleration of the ego-vehicle  \\ \hline
Curvature bounds &  $-\kappa_{max}\leq\vert\frac{\ddot{y}(t)\dot{x}(t)-\ddot{x}(t)\dot{y}(t)}{(\dot{x}(t)^2+\dot{y}(t)^2)^{1.5}}\vert\leq \kappa_{max}$ & \makecell{$-\kappa_{max}$: maximum curvature bound for the \\ ego-vehicle trajectory}  \\ \hline
Centripetal Acceleration bounds  & $-c_{max}\leq \dot{x}(t)^2\kappa(x(t)) \leq c_{max}$ & \makecell{$c_{max}$: maximum centripetal acceleration bound \\for the ego-vehicle.}  \\ \hline
Lane boundary & $y_{lb}(x(t))\leq y(t)\leq y_{ub}(x(t))$ & \makecell{$y_{lb}, y_{ub}$: Lane limits as a function \\of the ego-vehicle's position.} \\ \hline
\end{tabular}
\normalsize
\label{ineq_list}
\vspace{-0.6cm}
\end{table*}

\subsection{Reformulating Quadratic Inequalities}

\noindent The collision avoidance constraints can be re-written in the following form:

\small
\begin{align}
    \textbf{f}_{o, i} = \left \{ \begin{array}{lcr}
x(t) -x_{o, i}(t)-d_{o, i}(t)\cos\alpha_{o, i}(t) \\
y(t) -y_{o, i}(t)-d_{o, i}(t)\sin\alpha_{o, i}(t) \\ 
\end{array} \right \} d_{o, i}(t)\geq 1
\label{sphere_proposed}
\end{align}
\normalsize
\vspace{-0.1cm}
\noindent where $\alpha_{o, i}(t)$ represents the angle that the line-of-sight vector between the ego-vehicle and its $i^{th}$ neighbor makes with the $X$ axis. Similarly, the variable $d_{o, i}(t)$ represents the ratio of the length of this vector with the minimum distance separation required for collision avoidance. Following a similar approach, we can rephrase the velocity and acceleration bounds from Table \ref{ineq_list} as:

\vspace{-0.3cm}

\small
\begin{align}
    \textbf{f}_{v} = \left \{ \begin{array}{lcr}
\dot{x}(t) -d_{v}(t)\cos\alpha_{v}(t) \\
\dot{y}(t) -d_{v}(t)\sin\alpha_{v}(t)\\ 
\end{array} \right \}, v_{min}\leq d_{v}(t)\leq v_{max}
\label{vel_bound_proposed}
\end{align}
\normalsize

\vspace{-0.5cm}
\small
\begin{align}
    \textbf{f}_{a} = \left \{ \begin{array}{lcr}
\ddot{x}(t) -d_{a}(t)\cos\alpha_{a}(t) \\
\ddot{y}(t) -d_{a}(t)\sin\alpha_{a}(t)\\ 
\end{array} \right \}, 0\leq d_{a}(t)\leq a_{max}
\label{acc_bound_proposed}
\end{align}
\normalsize

The variables $\alpha_{o, i}(t)$, $\alpha_{o, i}(t)$, $\alpha_{a, i}(t)$, $d_{o, i}(t)$, $d_{v, i}(t)$, and $d_{a, i}(t)$  are additional variables that will be obtained by our batch projection optimizer along with $\boldsymbol{\xi}_j$.

Using \eqref{vel_bound_proposed}-\ref{acc_bound_proposed},  curvature and centripetal acceleration can also be reduced to the \eqref{curvature_bounds} and \eqref{cent_bounds} respectively.

\small
\begin{subequations}
\begin{align}
    \frac{d_a(t)\vert \sin(\alpha_a(t)-\alpha_v(t) \vert}{d_v(t)^2}\leq \kappa_{max}
    \label{curvature_bounds}\\
    {d}_v(t)^2 \cos\alpha_v(t)^2 \vert\kappa(x(t))\vert\leq c_{max} \label{cent_bounds}
\end{align}
\end{subequations}
\normalsize 

\noindent The variable $\kappa(x(t))$ represents the curvature of the road. We approximate it as the curvature of the reference reference center-line aligned with the road geometry. Note that $\kappa(x(t))$ depends on the ego-vehicles position $x(t)$ with respect to the center-line.

\subsubsection{Reformulated Problem} Using the developments in the previous section and the trajectory parametrization presented in \eqref{param}, we can now replace the projection operator on line 8 of Alg. \ref{algo_1} with the following. Note that \eqref{lane_reform} is the matrix representation of the lane boundary constraints presented in Table \ref{ineq_list}.

\small
\begin{subequations}
\begin{align}
    \boldsymbol{\xi}_j^{*} = \arg\min_{{\boldsymbol{\xi}}_j}\frac{1}{2}\Vert \boldsymbol{\xi}_j-\overline{\boldsymbol{\xi}}_j\Vert_2^2\label{cost_reform}  \\
    \textbf{A}_{eq} {\boldsymbol{\xi}}_j= \textbf{b}_{eq} \label{eq_reform} \\
    \textbf{F} \boldsymbol{\xi}_j = \textbf{h}_j(\boldsymbol{\alpha}_j, \textbf{d}_j) \label{nonconvex_reform}  \\
    \textbf{d}_{o, j} \geq 1, \hspace{0.1cm} v_{min}\leq \textbf{d}_{v,j} \leq v_{max}, \hspace{0.1cm} 0\leq \textbf{d}_{a, j} \leq a_{max} \label{d_reform_1}\\
    \textbf{d}_{a, j}\vert\sin(\boldsymbol{\alpha}_{a, j}-\boldsymbol{\alpha}_{v, j})\vert\leq \textbf{d}_{v, j}^2\kappa_{max} \label{d_reform_2}\\
    (\textbf{d}_{v, j}\cos\boldsymbol{\alpha}_{v, j})^2\kappa(\boldsymbol{\xi}_j)  \leq c_{max}\label{d_reform_3}.\\
    \textbf{G}\boldsymbol{\xi}_j \leq \textbf{e}(\boldsymbol{\xi}_j) \label{lane_reform}
\end{align}
\end{subequations}
\normalsize

\small
\begin{align}
    \textbf{F} = \begin{bmatrix}
    \begin{bmatrix}
    \textbf{F}_{o}\\
    \dot{\textbf{W}}\\
    \ddot{\textbf{W}}
    \end{bmatrix} & \textbf{0}\\
    \textbf{0} & \begin{bmatrix}
    \textbf{F}_{o}\\
    \dot{\textbf{W}}\\
    \ddot{\textbf{W}}
    \end{bmatrix} 
    \end{bmatrix}, \textbf{h}_j = \begin{bmatrix}
    \textbf{x}_o+a \textbf{d}_{o, j}\cos\boldsymbol{\alpha}_{o, j}\\
     \textbf{d}_{v, j}\cos\boldsymbol{\alpha}_{v, j}\\
  \textbf{d}_{a, j}\cos\boldsymbol{\alpha}_{a, j}\\
 \textbf{x}_o+a \textbf{d}_{o, j}\sin\boldsymbol{\alpha}_{o, j}\\
     \textbf{d}_{v, j}\sin\boldsymbol{\alpha}_{v, j}\\
  \textbf{d}_{a, j}\sin\boldsymbol{\alpha}_{a, j}\\
    \end{bmatrix},
\end{align}
\normalsize

\begin{align*}
    \boldsymbol{\alpha}_j = (\boldsymbol{\alpha}_{o, j}, \boldsymbol{\alpha}_{a,j}, \boldsymbol{\alpha}_{v,j}), \qquad \textbf{d}_j =  (\textbf{d}_{o, j}, \textbf{d}_{v, j}, \textbf{d}_{a, j})
\end{align*}

The matrix $\textbf{F}_o$ is obtained by stacking the matrix $\textbf{W}$ from (\ref{param}) as many times as the number of neighboring vehicles considered for collision avoidance at a given planning cycle. The vector $\textbf{x}_o, \textbf{y}_o$ is formed by appropriately stacking $x_{o, i}(t), y_{o, i}(t)$ at different time instants and for all the neighbors. Similar construction is followed to obtain $\boldsymbol{\alpha}_{o}, \boldsymbol{\alpha}_{v}, \boldsymbol{\alpha}_{a}, \textbf{d}_{o}, \boldsymbol{d}_{v} \boldsymbol{d}_{a}$.

\noindent Constraints \eqref{nonconvex_reform}-\eqref{lane_reform} acts as substitutes for $\textbf{g}(\boldsymbol{\xi}_j)\leq 0 $ in the projection operator (line 8, Alg.\ref{algo_1}). Please also note the addition of subscript $j$ indicating that the \eqref{cost_reform}-\eqref{lane_reform} is defined for the $j^{th}$ sample of ${\boldsymbol{\xi}}_j$

The approach of \cite{masnavi2022visibility} can efficiently handle \eqref{cost_reform}-\eqref{d_reform_1}. Thus, our aim is to accommodate the  new constraints \eqref{d_reform_2}-\eqref{lane_reform} while incurring minimal change to the batch update rule proposed in \cite{masnavi2022visibility}. This motivates our solution process that is discussed next.

\subsubsection{Solution Process} The projection process relies on two key ideas. First, we relax the non-convex equality \eqref{nonconvex_reform} and affine inequality constraints as $l_2$ penalties and augment them into the projection cost \eqref{cost_reform}.

\small
\begin{dmath}
    \mathcal{L}(\boldsymbol{\xi}_j, \boldsymbol{\lambda}_j) = \frac{1}{2}\left\Vert \boldsymbol{\xi}_j-\overline{\boldsymbol{\xi}}_j\right\Vert_2^2-\langle \boldsymbol{\lambda}_{j}, {\boldsymbol{\xi}}_j\rangle+\frac{\rho}{2} \left \Vert \textbf{F} {\boldsymbol{\xi}}_j-\textbf{h}_j\right \Vert_2^2+  \frac{\rho}{2}\left \Vert \mathbf{G} \boldsymbol{\xi}_{j} - \textbf{e}(\boldsymbol{\xi}_j) + \mathbf{s}_j \right \Vert^2
    \label{aug_lag}
\end{dmath}
\normalsize

\begin{algorithm*}[!h]
\centering
 \caption{Efficient Batch Optimization for Lower Level Trajectory Planning}\label{algo_batch}
 \small
    \begin{algorithmic}[1]   
    \State Initialize ${^k}\boldsymbol{\lambda}_{j}, {^k}\boldsymbol{\xi}_j, {^k}\boldsymbol{\alpha}_{o, j}, {^k}\boldsymbol{\alpha}_{v, j}, {^k}\boldsymbol{\alpha}_{a, j} $, ${^k}\textbf{d}_{o,j}$, ${^k}\textbf{d}_{v,j}$, ${^k}\textbf{d}_{a,j}$ for iteration $k=0$\\
    \For{$k=1, k \leq \text{maxiter}, k++$}
    {     
    \small
    \begin{align}
        {^{k+1}}{\boldsymbol{\xi}}_j = \arg\min_{{\boldsymbol{\xi}}_j} 
            \frac{1}{2}\left\Vert \boldsymbol{\xi}_j-\overline{\boldsymbol{\xi}}_j\right\Vert_2^2-\langle {^k}\boldsymbol{\lambda}_{j}, {\boldsymbol{\xi}}_j\rangle+\frac{\rho}{2}\left\Vert \textbf{F}{\boldsymbol{\xi}}_j  -\textbf{h}_j({^k}\boldsymbol{\alpha}_j, {^k}\textbf{d}_j ) \right \Vert_2^2+\frac{\rho}{2}\left \Vert \mathbf{G} \boldsymbol{\xi}_j - \mathbf{e}({^k}\boldsymbol{\xi}_j) + {^{k}}\mathbf{s}_{j} \right \Vert^2 \nonumber \\,
            \textbf{A}_{eq}{\boldsymbol{\xi}}_j  =\textbf{b}_{eq}  
             \label{split_xy}
    \end{align}
    \normalsize
    
    \small
    \begin{align}
        {^{k+1}}\boldsymbol{\alpha}_{j}, {^{k+1}}\textbf{d}_{j}  = \arg\min_{\boldsymbol{\alpha}_{j}, \textbf{d}_j}\left\Vert \textbf{F}{^{k+1}}{\boldsymbol{\xi}}_j-\textbf{h}_j(\boldsymbol{\alpha}_{j}, \textbf{d}_{j})\right\Vert_2^2
        \label{split_alpha_o}
    \end{align}
    \normalsize
    
    \small
    \begin{subequations}
    \begin{align}
        {^{k+1}}\textbf{d}_{o, j} = clip(1, \infty)
        \label{clip_d_o}
    \end{align}
    \end{subequations}
    \normalsize
    \small
    \begin{subequations}
    \begin{align}
             {^{k+1}}\textbf{d}_{v, j} = clip(\overline{v}_{min}, \overline{v}_{max}) \label{clip_d_v}\\
            \overline{v}_{min} = \max(v_{min}, \sqrt{{^k}\textbf{d}_{a, j}\vert\sin({^{k+1}}\boldsymbol{\alpha}_{a, j}-{^{k+1}}\boldsymbol{\alpha}_{v, j}\vert}  )    )\\
        \overline{v}_{max} = \min(v_{max}, \sqrt{c_{max}/(\vert\kappa({^k}\boldsymbol{\xi}_j)\vert(\cos\boldsymbol{{^{k+1}}\alpha}_{v, j})^2)}
    \end{align}
    \end{subequations}
    \normalsize
    \small
    \begin{subequations}
    \begin{align}
             {^{k+1}}\textbf{d}_{a, j} = clip(0, \overline{a}_{max}) \label{clip_d_a}\\
            \overline{a}_{max} =  \min \left( a_{max},  \frac{{^{k+1}}\textbf{d}_v^2 \kappa_{max} }{\vert\sin(\boldsymbol{{^{k+1}}\alpha}_{a, j}- \boldsymbol{{^{k+1}}\alpha}_{v, j} )\vert } \right)
    \end{align}
    \end{subequations}
    \normalsize
    
    \small
    \begin{subequations}
    \begin{align}
         {^{k+1}}\mathbf{s} =& \text{max}\left(0, -\mathbf{G} {^{k+1}}\boldsymbol{\xi}_{j} - \mathbf{e}({^{k+1}}\boldsymbol{\xi}_j)\right) \label{update_slack}\\
            {^{k+1}}\boldsymbol{\lambda} =& {^{k}}\boldsymbol{\lambda} - \frac{\rho}{2} \mathbf{F}^T\left( \mathbf{F} {^{k+1}}\boldsymbol{\xi}_{j} - \mathbf{h}({^{k+1}}\boldsymbol{\alpha}_{j}, {^{k+1}}\textbf{d}_{j})\right)  \frac{\rho}{2}\mathbf{G}^T\left(\mathbf{G} {^{k+1}}\boldsymbol{\xi}_{j} - \mathbf{e}({^{k+1}}\boldsymbol{\xi}_j) + {^{k+1}}\mathbf{s}\right) \label{update_lagrange}
    \end{align}
    \end{subequations}
    \normalsize
    } 
\State Return ${^{k+1}}{\boldsymbol{\xi}}_j$
\end{algorithmic}
\end{algorithm*}

\noindent The constant $\rho$ controls the trade-off between minimizing the projection cost and constraint residuals. The slack variable $\textbf{s}_j\geq \textbf{0}$ is unknown and will computed by the projection process along with $\boldsymbol{\xi}_j^{*}$ and other variables. The variable $\boldsymbol{\lambda}_j$ is the so-called  Lagrange multipliers and play a crucial role in driving the residuals to zero for any arbitrary $\rho$ \cite{admm_neural}. 

The second part of the idea is to apply Alternating Minimization (AM) approach for computing the optimal solution of \eqref{aug_lag} subject to \eqref{nonconvex_reform}-\eqref{lane_reform}. This is presented in Algorithm \ref{algo_batch}. At each step, we minimize only one block of variables while others are fixed at values obtained in the preceding iteration or the previous step of the current iteration. We elaborate on each step next.

\subsubsection{Analysis \textbf{Step} \eqref{split_xy} } At step $k+1$, we use the ${^k}\boldsymbol{\xi}_j$ to make an estimate of the exact lane bounds $\textbf{e}({^k}\boldsymbol{\xi}_j)$ and curvature $\kappa({^k}\boldsymbol{\xi}_j)$. As a result, the curvature and lane bounds become independent of $\boldsymbol{\xi}_j$ and \ref{split_xy} turns to a equality constrained QP of the form presented in \eqref{over_2}. The solution for the entire batch can computed in one-shot using \eqref{over_3}

\subsubsection{Analysis \textbf{Step} \eqref{split_alpha_o} } For a given ${^{k+1}}\boldsymbol{\xi}_j$, the optimization over $\boldsymbol{\alpha}_j, \textbf{d}_j$ have a closed form solution that can be evaluated across all the batch in one-shot \cite{masnavi2022visibility} . In this step, we ignore the constraints on $\boldsymbol{\alpha}_j, \textbf{d}_j$.

\subsubsection{Analysis of \textbf{Step} \eqref{clip_d_o}-\eqref{clip_d_a} } In these steps, we clip the values of $\textbf{d}_{o, j}, \textbf{d}_{v, j}, \textbf{d}_{a, j}$ computed in the previous step to satisfy their respective bounds. The clipping over $\textbf{d}_{o, j}$ follows from \cite{masnavi2022visibility} while the remaining two are new additions which are obtained in the following manner. For a given ${^{k+1}}\boldsymbol{\alpha}_{v, j}, {^{k+1}}\boldsymbol{\alpha}_{a, j}$ obtained in \eqref{split_alpha_o} and ${^k}\textbf{d}_{a, j}$ from previous iteration, the curvature \eqref{d_reform_2} and centripetal acceleration \eqref{d_reform_3} constraints  reduce to just upper and lower bounds on $\textbf{d}_{v, j}$. Similar reasoning leads to \eqref{clip_d_a}.

\subsubsection{Analysis of \textbf{Step} } The update \eqref{update_slack}-\eqref{update_lagrange} follows from \cite{ghadimi2014optimal}, \cite{admm_neural} and herein, we update the Lagrange multipliers and slack variables based on the residuals at the current iteration.

\noindent \textbf{Summary:} Using a combination of clever reformulations and a AM approach, each step of the batch projection either boils down to (i) batch QP (\eqref{split_xy}) or simply function evaluations (\eqref{split_alpha_o}-\eqref{update_lagrange}  ).

\end{document}